\documentclass[runningheads]{llncs}
\usepackage{graphicx}
\usepackage{authblk}
\usepackage{graphicx}
\usepackage{subfigure}
\usepackage{array}
\usepackage{float}
\usepackage{booktabs}
\usepackage{bbding}
\usepackage{etoolbox}
\usepackage{color}
\usepackage{soul}
\usepackage{algorithm}
\usepackage{algorithmic}
\usepackage{cite}
\usepackage{balance}
\usepackage{amsmath, amssymb}
\usepackage{makecell}
\usepackage{bm}
\usepackage{multirow}
\usepackage{color,xcolor}
\usepackage{hyperref}

\hypersetup{
    colorlinks,
    linkcolor={blue!50!black},
    citecolor={blue!50!black},
    urlcolor={blue!50!black}
}

\begin{document}
%
\title{Endo-4DGS: Endoscopic Monocular Scene Reconstruction with 4D Gaussian Splatting}

%
\titlerunning{Endo-4DGS}
\authorrunning{Y. Huang et al.}
\author{Yiming Huang\inst{1~\star}
\and Beilei Cui\inst{1~\star}
\and Long Bai\inst{1}
\thanks{Co-first authors.}
\and Ziqi Guo\inst{1}
\and Mengya Xu\inst{1}
\and Mobarakol Islam\inst{3}
\and Hongliang Ren\inst{1,2,4} 
\thanks{Corresponding author.}}
\institute{Department of Electronic Engineering, The Chinese University of Hong Kong (CUHK), Hong Kong SAR, China
\and Shun Hing Institute of Advanced Engineering, CUHK, Hong Kong SAR, China \and Wellcome/EPSRC Centre for Interventional and Surgical Sciences (WEISS), University College London, UK
\and Shenzhen Research Institute, CUHK, Shenzhen, China\\
\email{\{yhuangdl, beileicui, b.long, 1155199284\}@link.cuhk.edu.hk, mengya@u.nus.edu, mobarakol.islam@ucl.ac.uk, hlren@ee.cuhk.edu.hk}}

\maketitle              
\begin{abstract}

In the realm of robot-assisted minimally invasive surgery, dynamic scene reconstruction can significantly enhance downstream tasks and improve surgical outcomes. Neural Radiance Fields (NeRF)-based methods have recently risen to prominence for their exceptional ability to reconstruct scenes but are hampered by slow inference speed, prolonged training, and inconsistent depth estimation. Some previous work utilizes ground truth depth for optimization but is hard to acquire in the surgical domain. To overcome these obstacles, we present Endo-4DGS, a real-time endoscopic dynamic reconstruction approach that utilizes 3D Gaussian Splatting (GS) for 3D representation. Specifically, we propose lightweight MLPs to capture temporal dynamics with Gaussian deformation fields. To obtain a satisfactory Gaussian Initialization, we exploit a powerful depth estimation foundation model, Depth-Anything, to generate pseudo-depth maps as a geometry prior. We additionally propose confidence-guided learning to tackle the ill-pose problems in monocular depth estimation and enhance the depth-guided reconstruction with surface normal constraints and depth regularization. Our approach has been validated on two surgical datasets, where it can effectively render in real-time, compute efficiently, and reconstruct with remarkable accuracy. Our code is available at~\url{https://github.com/lastbasket/Endo-4DGS}.

\end{abstract}
\section{Introduction}
Endoscopic procedures have become a cornerstone in minimally invasive surgery, offering patients with reduced trauma and quicker recovery times~\cite{gao2023transendoscopic,psychogyios2023sar,zia2023surgical}. In this case, accurate and dynamic 3D reconstruction of the endoscopic scene is critical to enhancing the surgeon's spatial understanding and navigation, facilitating more precise and efficient interventions~\cite{mahmoud2017orbslam}. However, the complex and constrained nature of endoscopic scenes poses significant challenges for traditional 3D reconstruction techniques due to factors such as limited field-of-view, occlusions, and dynamic tissue deformation~\cite{wang2022neural,zha2023endosurf,yang2023neural}.

Recent advancements in endoscopic 3D reconstruction have been boosted by the capabilities of Deep Neural Networks (DNNs)~\cite{stucker2020resdepth} and Neural Radiance Fields (NeRFs)~\cite{mildenhall2021nerf}. Some studies have achieved strong performance in depth estimation and reconstruction under endoscopy, particularly through stereo reconstruction~\cite{bae2020deep,long2021dssr}, structure from motion~\cite{barbed2023tracking}, depth and pose estimation~\cite{shao2022self,ozyoruk2021endoslam} or extensive visual pre-training~\cite{beilei2024surgical}.  EndoNeRF~\cite{wang2022neural} is the first to leverage NeRF~\cite{mildenhall2021nerf} in endoscopic scenes by dual neural fields approach to model tissue deformation and canonical density. EndoSurf~\cite{zha2023endosurf} further employs signed distance functions to model tissue surfaces, imposing explicit self-consistency constraints on the neural field. To tackle the lengthy training time requirement, LerPlane~\cite{yang2023neural} constructs a 4D volume by introducing 1D time to the existing 3D spatial space. This extension allows for the formulation of both static fields and dynamic fields by utilizing the spatial-temporal planes, respectively, which leads to a substantial decrease in computational resources. However, reconstructing high-dimensional deformable scenes in real-time remains a challenge.

\begin{figure}[!t]
    \centering
    \includegraphics[width=0.9\linewidth]{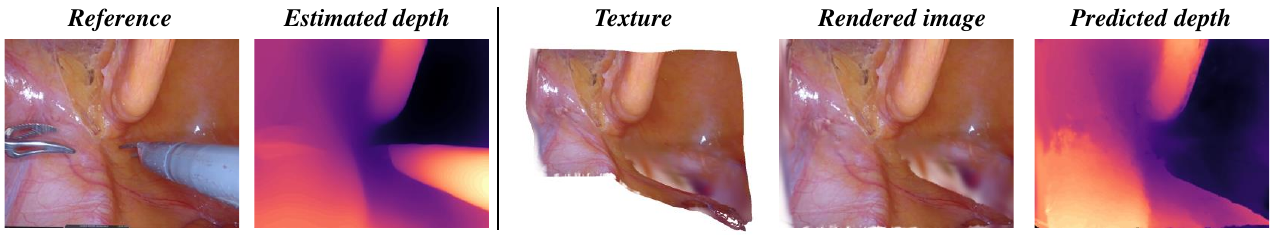}
    \caption{Ground truth reference, estimated depth from Depth-Anything; 3D textures, rendered image, and predicted depth of our proposed method.}
    \label{fig:intro}
\end{figure}

NeRF-based methods have revolutionized 3D scene reconstruction but face challenges such as slow rendering speeds and suboptimal localization accuracy~\cite{chen2024survey}. Addressing these issues, 3D Gaussian Splatting (GS) has emerged as an effective alternative, offering fast inference and superior 3D representation~\cite{kerbl20233d}. By optimizing anisotropic 3D Gaussians using a set of scene images, 3D GS successfully captures the spatial positioning, orientations, color properties, and alpha blending factors, reconstructing both the geometry and visual texture of the scene. The method's tile-based rasterizer further guarantees fast rendering performance. 

To tackle the deformable tissue reconstruction challenges in endoscopic scenes, we further incorporate the temporal dimension as the fourth axis to model dynamic environments~\cite{wu20234dgaussians}. Moreover, current solutions for depth prior-assisted reconstruction depend on multi-view information and the static scene assumption~\cite{wang2023sparsenerf, chung2023depth}, which are not always feasible in the surgical scenario. Meanwhile, the predictions of existing monocular depth estimation methods~\cite{depthanything} also suffer from ill-posed problems. The predicted depth results in uncertain measurements even with little changes in the environment, e.g. small deformation on the tissues. Therefore, reconstruction using depth prior supervision remains a challenge in deformable surgery scenarios. 
To overcome these hurdles, we leverage Depth-Anything~\cite{depthanything}, a cutting-edge method educated through extensive visual pre-training that has demonstrated remarkable depth estimation capabilities across various scenarios. By applying Depth-Anything, we project the pre-trained depth into 3D for more robust 4D Gaussian initialization. To address the challenges posed by inaccurate issues in estimating depth using a monocular camera, we introduce a confidence-guided learning approach that effectively reduces the influence of noisy or uncertain measurements in the pre-trained depth estimation. We additionally implement surface normal constraints and depth regularization to strengthen the pseudo-depth's accuracy and geometry constraint. Fig.~\ref{fig:intro} showcases our 3D textures, the rendered images, and the depth predictions for endoscopic views.
Specifically, our contributions in this paper are threefold:
\begin{itemize}
    \item We present Endo-4DGS, an innovative technique that adapts Gaussian Splatting for endoscopic scene reconstruction. Utilizing pseudo-depth generated by Depth-Anything, Endo-4DGS achieves remarkable reconstruction outcomes without needing ground truth depth data.
    \item We propose confidence-guided learning to tackle the ill-pose monocular depth adaption problems, and further employ depth regularization and surface normal constraints against the depth prior adaption challenge in the deformable surgical reconstruction task.
    \item Our extensive validation on two real surgical datasets shows that Endo-4DGS attains high-quality reconstruction, excels in real-time performance, reduces training expenditures, and demands less GPU memory, which sets the stage for advancements in robot-assisted surgery.
\end{itemize}

\section{Methodology}

In this section, we introduce the representation and rendering formula of 4D Gaussians~\cite{wu20234dgaussians} in Sec.~\ref{sec.2.1} and demonstrate our motivation and detailed implementation of the depth prior-based reconstruction in Sec.~\ref{sec.2.2}.

\subsection{Preliminaries}\label{sec.2.1} 
3D GS~\cite{kerbl20233d} utilizes 3D differentiable Gaussians as the unstructured representation, allowing for a differentiable volumetric representation that can be rapidly rasterized and projected onto a 2D surface for swift rendering. With covariance matrix ${\rm\mathbf{\Sigma}}$ and mean $\mu$ the 3D GS at position $x$ is described as $G(x)=e^{-\frac{1}{2}(x-\mu)^T{\rm \mathbf{\Sigma}}^{-1}(x-\mu)}$, where the covariance ${\rm \mathbf{\Sigma}}$ can be further decomposed into ${\rm\mathbf{\Sigma}} = \mathbf{R}\mathbf{S}\mathbf{S}^T\mathbf{R}^T$ with the scaling $\mathbf{S}$ and rotation $\mathbf{R}$. Introducing by ~\cite{yifan2019differentiable}, with the viewing transform $\mathbf{W}$ and the Jacobian of the affine approximation of the projective transformation $\mathbf{J}$, covariance in the camera plane can be described as ${\rm\mathbf{\Sigma}}^{\prime} = \mathbf{J}\mathbf{W}{\rm \mathbf{\Sigma}} \mathbf{W}^T\mathbf{J}^T$. The final rendering equation is:
\begin{equation}
     \hat{C} = \sum_{i\in N}c_i \alpha_i \prod_{j=1}^{i-1} (1-\alpha_i),
\end{equation}
\noindent where $\hat{C}$ is the predicted pixel color from $N$ points. $c_i$, $\alpha_i$ are the color defined by the spherical harmonics coefficients and the density calculated by multiplying the 2D covariance $\mathbf{\Sigma}^\prime$ with the learned opacity $o_i$.
\subsection{Proposed Methodology}\label{sec.2.2}

\begin{figure}[!t]
    \centering
    \includegraphics[width=0.95\linewidth]{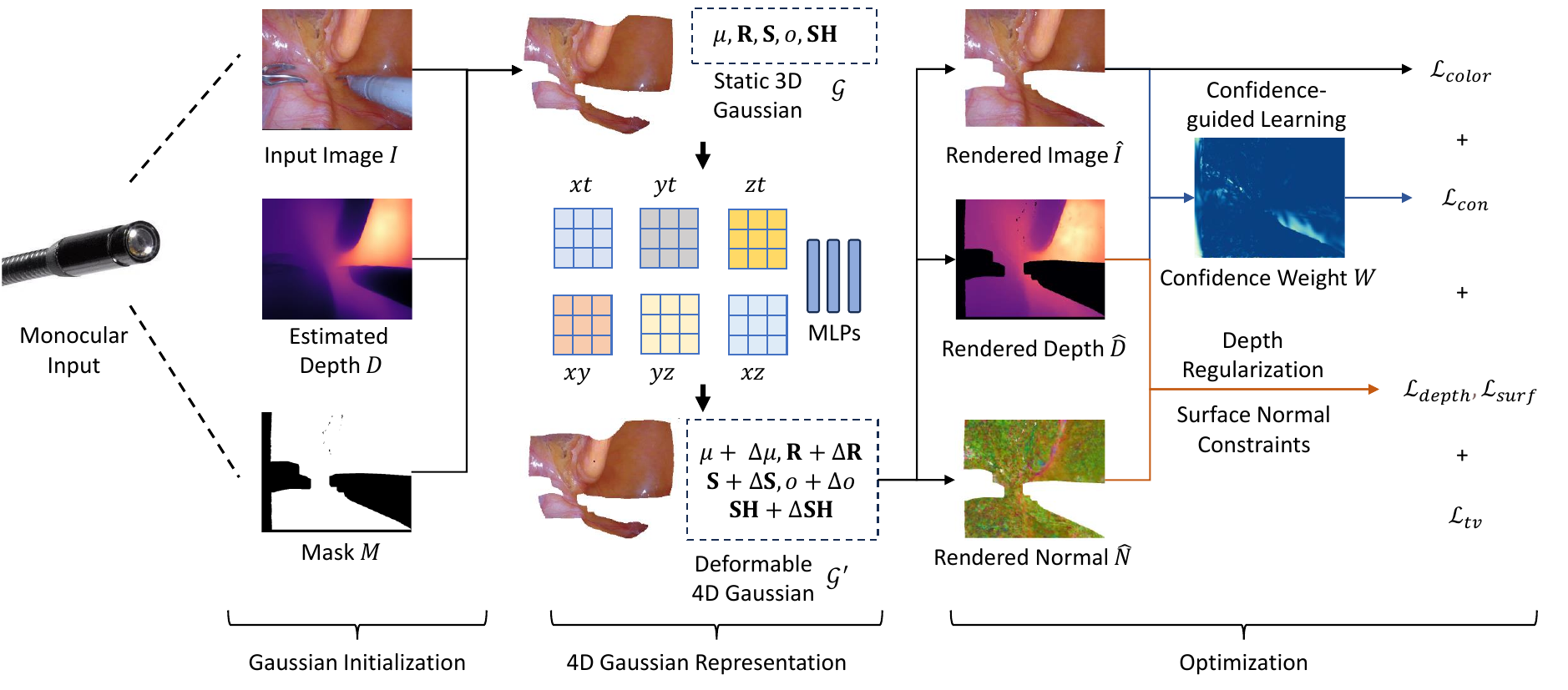}
    \caption{Illustration of our proposed Endo-4DGS framework. We utilize monocular images, estimated depths from Depth-Anything, and surgical tool masks for training. 3D Guassian is represented as $\mathcal{G}$ with position mean $\mu$, rotation $\mathbf{R}$, scaling $\mathbf{S}$ opacity $o$, and spherical harmonics $\mathbf{SH}$. 4D Gaussian is described as $\mathcal{G}^\prime=\mathcal{G}+\Delta\mathcal{G}$. $\mathcal{L}_{color}, \mathcal{L}_{con}, \mathcal{L}_{depth}, \mathcal{L}_{surf}, \mathcal{L}_{tv}$ are the color loss, confidence loss, depth regularization loss, surface normal loss and total-variational loss, respectively.}
    \label{fig:main}
\end{figure}

\noindent \textbf{4D Gaussian Splatting for Deformable Scene Representation}
Inspired by~\cite{wu20234dgaussians}, we represent the deformable surgical scene with the 4D Gaussian $\mathcal{G}^\prime=\Delta \mathcal{G} + \mathcal{G}$ which includes a static 3D Gaussian $\mathcal{G}$ and its deformation $\Delta \mathcal{G}=\mathcal{F}(\mathcal{G}, t)$, where $\mathcal{F}$ is the deformation network and $t$ is the time. The spatial-temporal encoder $\mathcal{H}$ is defined with multi-resolution Hexplanes $R_l(i,j)$ and a tiny MLP $\phi_d$,  $\mathcal{H}(\mathcal{G},t)=\{R_l(i,j), \phi_d | (i,j) \in \{(x,y),(x,z),(y,z), (x,t),(y,t),(z,t)\}, l \in \{1,2\}\}$, and the spatial-temporal feature is encoded as $f_d=\mathcal{H}(\mathcal{G}, t)$.

A multi-head Gaussian deformation decoder $\mathcal{D}=\{\phi_\mu, \phi_r, \phi_s, \phi_o, \phi_\mathbf{SH}\}$ is designed for decoding the deformation of position, rotation, scaling, opacity and spherical harmonics $\mathbf{SH}$ with five tiny MLPs. The final representation of 4D Gaussian can be expressed as:
\begin{equation}
    \begin{aligned}
        \mathcal{G}^\prime &= \{\mu+\phi_\mu(f_d), r+\phi_r(f_d), s+\phi_s(f_d), o+\phi_o(f_d), \mathbf{SH}+\phi_{\mathbf{SH}}(f_d)\}\\
        &=\{\mu+\Delta\mu, \mathbf{R}+\Delta \mathbf{R}, \mathbf{S}+\Delta \mathbf{S}, o+\Delta o, \mathbf{SH}+\Delta \mathbf{SH}\}
    \end{aligned}
\end{equation}

\noindent \textbf{Gaussians Initialization with Depth Prior.}
Retrieving accurate point clouds in surgical scenes is challenging since there is only monocular visual information from the consumer-level endoscopes. Therefore, we propose to use the pre-trained depth to implement the point cloud initialization for the 4D Gaussian. With the pre-trained depth estimation model and the input image $I$, we estimate an inverse depth map $D_{inv}$. Then a scaling $\beta$ is applied to recover the depth map $D = \frac{\beta}{D_{inv}}$ in the camera coordinate. Given the camera intrinsic matrix $K_1$, and the extrinsic matrix $K_2$, we project the point cloud $P\in \mathbb{R}^{N\times 3}$ with size $N$ from the given image $I$ as follows:

\begin{equation}
    P = K_2^{-1}K_1^{-1}[(I \odot M), D],
\end{equation}
where $M$ is the mask for the input image, $\odot$ is the element-wise multiplication, and $[\cdot]$ indicates concatenation. With the point cloud from depth prior, We initialize $\mu, \mathbf{R}$, making the training process faster for convergence and more robust in terms of geometry.
\\
\\
\noindent \textbf{Confidence Guided Learning.} Monocular reconstruction with estimated depth is an ill-pose problem since there is no access to the ground truth geometry information. Inspired by~\cite{wang2024masked, chung2023depth}, we formulate our solution with a probabilistic model to learn statistics for depth from Depth-Anything, which is defined as:
\begin{equation}
     \hat{D} = \frac{\sum_{i\in N}d_i \alpha_i \prod_{j=1}^{i-1} (1-\alpha_i)}{\sum_{i\in N}W_i}, \ W_i = \alpha_i \prod_{j=1}^{i-1} (1-\alpha_i)
\end{equation}

\noindent where $d_i$ is the depth of the center of the Gaussian obtained by projecting to the z-axis of the camera coordinate. $W_i\in (0, 1)$ is defined as the confidence weight for the corresponding point, which is closer to 1 with higher confidence. Following the above definition, the confidence guidance loss can be expressed as:
\begin{equation}
    \mathcal{L}_{con} = \mathbb{E}[\frac{1}{2W^2}||\hat{D}_{norm}-D_{norm}||_2^2 + \log(W)] + \mathbb{E}[\frac{1}{2W^2}||\hat{C}-C||_2^2 + \log(W)],
\end{equation}

\noindent where $\mathbb{E}(\cdot)$ is the expectation, $D_{norm}$ and $\hat{D}_{norm}$ are the depth prior and rendered depth normalized to $(0,1)$. While we penalize the depth and color with less confidence, we also add the $\log(\cdot)$ as a regularization term. The confidence weight, therefore, maximizes the error where the rendered depth is different from the depth prior while reducing the influence of the uncertain value of the pre-trained depth estimation. 
\\

\noindent \textbf{Surface Normal Constraints and Depth Regularization.}
To utilize the pre-trained depth map more effectively as the pseudo-ground truth, we propose to utilize depth regularization loss and surface normal loss. Following~\cite{cheng2023gaussian}, we approximate the surface normal $\hat{n}_i\in \hat{\mathbf{N}}$ with the shortest axis: 
\begin{equation}
    \mathbf{\hat{n}_i} = \mathbf{R}_i[r:], \ r=\text{argmin}([s_1, s_2, s_3]),
\end{equation}
where r is the index of the shortest scaling in $\mathbf{S}_i=diag(s_1, s_2, s_3)$ selected by $\text{argmin}(\cdot)$. Then we calculate the gradient of the depth prior $\nabla D = (G^W, G^H)$, and formulate the pseudo surface normal as:
\begin{equation}
    \mathbf{n}_i = [\frac{G^W_i}{\sqrt{(G^W_i)^2+(G^H_i)^2+1}}, \frac{G^H_i}{\sqrt{(G^W_i)^2+(G^H_i)^2+1}}, \frac{1}{\sqrt{(G^W_i)^2+(G^H_i)^2+1}}],
\end{equation} where $G^W, G^H$ are the gradients along the width and height of the depth map. The surface normal constraints is described as $\mathcal{L}_{surf} = \Vert \mathbf{N}-\hat{\mathbf{N}} \Vert_1$. We also regularize the predicted depth from 4D Gaussian with a normalized depth loss and gradient loss. The depth regularization term $\mathcal{L}_{depth}$ is expressed as:
\begin{align}
    \mathcal{L}_{depth} &= \lambda_{norm}\Vert D_{norm}-\hat{D}_{norm} \Vert_1 + \lambda_{grad}(1-P_{corr}(\Vert \nabla D \Vert_2, \Vert \nabla \hat{D} \Vert_2)),
\end{align}
\noindent where $P_{corr}(\cdot)$ is the Pearson Correlation Coefficient, $\lambda_{norm}, \lambda_{grad}$ are the weights for the normalized depth loss and gradient loss. 

With the $\mathcal{L}_{color}$ color loss and a grid-based total-variational loss $\mathcal{L}_{tv}$~\cite{cao2023hexplane, fang2022fast, kerbl20233d}, our final loss for optimizing can be represented as:
\begin{equation}
        \mathcal{L} = \mathcal{L}_{color} + \mathcal{L}_{tv} + \mathcal{L}_{depth} + \lambda_{surf}\mathcal{L}_{surf}+\lambda_{con} \mathcal{L}_{con},
\end{equation}
where $ \lambda_{surf}, \lambda_{con}$ are the weights for the surface constraints and confidence loss. Following~\cite{wu20234dgaussians}, we emit $\mathcal{L}_{tv}$ for the training of the static 3D Gaussians.

\section{Experiments}
\label{sec:exper}

\subsection{Dataset}
\label{sec:dataset}

We evaluate the performance based on two publicly available datasets, StereoMIS~\cite{hayoz2023robust} and EndoNeRF~\cite{wang2022neural}. The StereoMIS dataset~\cite{hayoz2023robust} is a stereo video dataset captured by the da Vinci Xi surgical system, consisting of 11 surgical sequences by in-vivo porcine subjects, where we extract the 800 to 1000 frames from the first scene. The EndoNeRF dataset~\cite{wang2022neural} includes two samples of prostatectomy via stereo cameras and provides estimated depth maps based on stereo-matching techniques, they also include challenging scenes with tool occlusion and non-rigid deformation. The training and validation splitting follows the 7:1 strategy in~\cite{zha2023endosurf}. We use PSNR, SSIM, and LPIPS to evaluate the 3D scene reconstruction performance. We also report the results of training time, inference speed, and GPU memory usage on one single RTX4090 GPU.

\subsection{Implementation Details}
\label{sec:implementation}

All experiments are conducted on the RTX4090 GPU with the Python PyTorch framework. We adopt the Adam optimizer with an initial learning rate of $1.6\times 10^{-3}$. We employ the Depth-Anything-Small model for pseudo-depth map generation with depth scale $\beta = 1000$ and $\lambda_{norm}=0.01, \lambda_{grad}=0.001, \lambda_{surf}=0.001, \lambda_{con}=0.0001$. We use an encoding voxel size of $[64, 64, 64, 75]$, where the four dimensions are length, width, height, and time, respectively.

\subsection{Results}
\label{sec:results}

\begin{table}[!h]
\caption{Comparison experiments on the EndoNeRF dataset~\cite{wang2022neural} against EndoNeRF~\cite{wang2022neural}, EndoSurf~\cite{zha2023endosurf}, and LerPlane~\cite{yang2023neural}. The best results are in bold.}
\centering
\resizebox{\textwidth}{!}{
\begin{tabular}{c|ccc|ccc|ccc}
\noalign{\smallskip}\hline
\multirow{2}{*}{Models}
& \multicolumn{3}{c|}{EndoNeRF-Cutting} & \multicolumn{3}{c|}{EndoNeRF-Pulling} & \multirow{2}{*}{\makecell[c]{Training \\ Time $\downarrow$}} & \multirow{2}{*}{FPS $\uparrow$} & \multirow{2}{*}{\makecell[c]{GPU \\ Usage $\uparrow$}}\\ \cline{2-7}
& PSNR $\uparrow$ & SSIM $\uparrow$ & LPIPS $\downarrow$ & PSNR $\uparrow$ & SSIM $\uparrow$ & LPIPS $\downarrow$ \\ \hline
EndoNeRF~\cite{wang2022neural} & 35.84 & 0.942 & 0.057 & 35.43 & 0.939 & 0.064 & 6 hours & 0.2 & 4 GB\\
EndoSurf~\cite{zha2023endosurf} & 34.89 & 0.952 & 0.107 & 34.91 & 0.955 & 0.120 & 7 hours & 0.04 & 17 GB\\
LerPlane-32k~\cite{yang2023neural} & 34.66 & 0.923 & 0.071 & 31.77 & 0.910 & 0.071 & 8 mins & 1.5 & 20 GB\\
\textbf{Endo-4DGS} & \textbf{36.56} & \textbf{0.955} & \textbf{0.032} & \textbf{37.85} & \textbf{0.959} & \textbf{0.043}  & \textbf{4 mins} & \textbf{100} & \textbf{4GB} \\\hline
\end{tabular}}
\label{tab:endonerf}
\end{table}


\begin{table}[!h]
\caption{Comparison experiments on the StereoMIS~\cite{hayoz2023robust}, against EndoNeRF~\cite{wang2022neural}, EndoSurf~\cite{zha2023endosurf}, and LerPlane~\cite{yang2023neural}. The best results are in bold.}
\centering
\resizebox{0.7\textwidth}{!}{
\begin{tabular}{c|ccc|ccc}
\noalign{\smallskip}\hline \multirow{2}{*}{Models} & PSNR $\uparrow$ & SSIM $\uparrow$ & LPIPS $\downarrow$ & \multicolumn{1}{c}{\makecell[c]{Training \\ Time $\downarrow$}} & FPS $\uparrow$ & \multicolumn{1}{c}{\makecell[c]{GPU \\ Usage $\downarrow$}}  \\ \hline
EndoNeRF~\cite{wang2022neural} & 21.49 & 0.622 & 0.360 & 5 hours & 0.2 &  4 GB\\
EndoSurf~\cite{zha2023endosurf} & 29.87 & 0.809 & 0.303& 8 hours & 0.04 &  14 GB \\
LerPlane-32k~\cite{yang2023neural} & 30.80 & 0.826 & 0.174 & 7 mins & 1.7 & 19 GB\\ 
\textbf{Endo-4DGS} & \textbf{32.69} & \textbf{0.850} & \textbf{0.148}& \textbf{7 mins} & \textbf{100} & \textbf{4 GB}\\\hline
\end{tabular}}
\label{tab:StereoMIS}
\end{table}

\begin{table}[t]
\centering
\caption{
    Ablation experiments of the proposed method on EndoNeRF dataset~\cite{wang2022neural}. To observe the performance changes, we remove (i) the depth regularization, (ii) the surface constraints, and (iii) the confidence guidance.  The best results are in bold.
}
\centering
\resizebox{0.8\textwidth}{!}{
 \label{tab:ablation}
\begin{tabular}{c|c|c|p{1.4cm}<{\centering} p{1.4cm}<{\centering} p{1.4cm}<{\centering}|p{1.4cm}<{\centering} p{1.4cm}<{\centering} p{1.4cm}<{\centering}}
\hline
\multirow{2}{*}{\makecell[c]{Depth\\Regularization }} & \multirow{2}{*}{\makecell[c]{Surface\\Constraints}} & \multirow{2}{*}{\makecell[c]{Confidence\\Guidance}} & 
\multicolumn{3}{c|}{EndoNeRF-Cutting} & \multicolumn{3}{c}{EndoNeRF-Pulling} \\ \cline{4-9}
& & & PSNR $\uparrow$ & SSIM $\uparrow$  & LPIPS $\downarrow$ & PSNR $\uparrow$ & SSIM $\uparrow$  & LPIPS $\downarrow$ \\ \hline
\XSolidBrush      &\XSolidBrush      & \multicolumn{1}{c|}{\XSolidBrush}         & 35.14 & 0.938 &  0.046 & 35.39 & 0.937 & 0.082 \\
\Checkmark      &\XSolidBrush      & \multicolumn{1}{c|}{\XSolidBrush}         & 36.00 & 0.949 & 0.040 & 35.68 &  0.942 & 0.072 \\
\XSolidBrush       & \Checkmark     & \multicolumn{1}{c|}{\XSolidBrush}         & 35.22 & 0.940 & 0.057 & 35.97 & 0.945 & 0.066\\
 \XSolidBrush      &\XSolidBrush      & \multicolumn{1}{c|}{\Checkmark}         & 35.54 & 0.941 & 0.048 & 35.68 & 0.942 & 0.066\\
\Checkmark       & \Checkmark     & \multicolumn{1}{c|}{\XSolidBrush}         & 36.24 & 0.951 & 0.038 & 36.35 & 0.945 & 0.062\\
\Checkmark      &\XSolidBrush      & \multicolumn{1}{c|}{\Checkmark}         & 36.22 & 0.951 & 0.036  & 36.94 & 0.952 & 0.053\\
\XSolidBrush       & \Checkmark    & \multicolumn{1}{c|}{\Checkmark}         & 36.08 & 0.946 & 0.036 & 36.15 & 0.943 & 0.064\\
\Checkmark     & \Checkmark     & \multicolumn{1}{c|}{\Checkmark}         & \textbf{36.56} & \textbf{0.955} & \textbf{0.032} & \textbf{37.85} & \textbf{0.959} & \textbf{0.043} \\ \hline
\end{tabular}}
\end{table}

We conducted a comprehensive comparison of our proposed method with state-of-the-art approaches for surgical scene reconstruction. Specifically, we reproduce EndoNeRF~\cite{wang2022neural}, EndoSurf~\cite{zha2023endosurf}, and LerPlane~\cite{yang2023neural} with the original implementation. The evaluation results on the EndoNeRF~\cite{wang2022neural} and StereoMIS~\cite{hayoz2023robust} datasets are presented in Table~\ref{tab:endonerf} and Table~\ref{tab:StereoMIS}. Upon analysis, we observed that while EndoNeRF~\cite{wang2022neural} and EndoSurf~\cite{zha2023endosurf} achieved relatively high performance, they required hours of training, making them time-consuming. On the other hand, LerPlane~\cite{yang2023neural} significantly reduced the training time to approximately 8 minutes but incurred a slight degradation in rendering performance. It is important to note that all of these state-of-the-art methods suffered from very low frames per second (FPS), which limited their practical application in real-time surgical scene reconstruction tasks. In contrast, our proposed method not only outperformed all evaluated metrics on both datasets but also achieved a real-time inference speed of 100 FPS, where the training was accomplished with only 4 minutes and 4GB of GPU memory. The significant improvement in inference speed makes our method highly suitable for real-time endoscopic applications.

\begin{figure}[!t]
    \centering
    \includegraphics[width=0.85\linewidth]{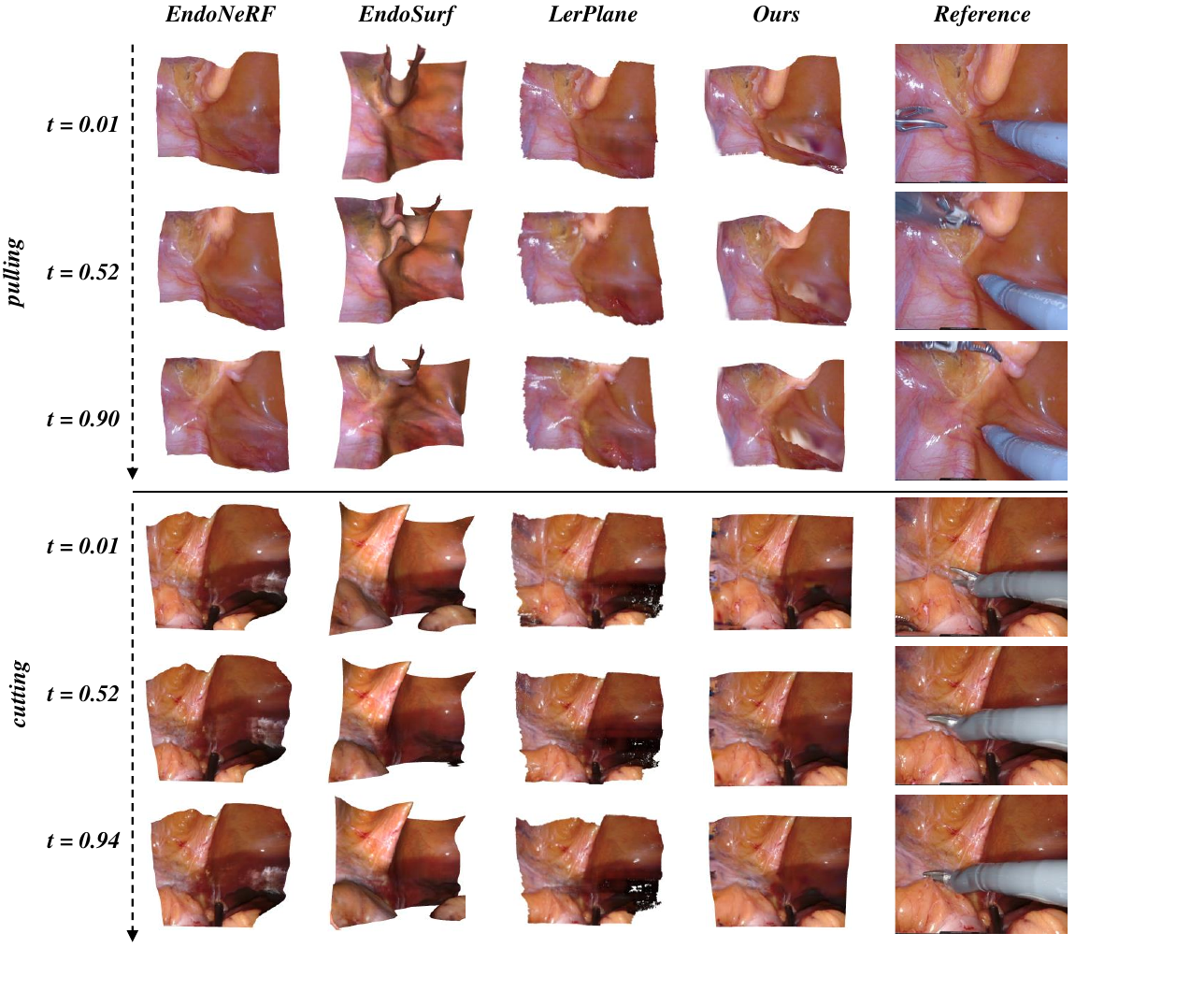}
    \caption{Qualitative comparison on the EndoNeRF dataset~\cite{wang2022neural} against EndoNeRF~\cite{wang2022neural}, EndoSurf~\cite{zha2023endosurf}, and LerPlane~\cite{yang2023neural}.}
    \label{fig:endonerf}
\end{figure}

We have provided qualitative results for EndoNeRF datasets~\cite{wang2022neural} in Fig.~\ref{fig:endonerf}. Notably, the visualizations demonstrate that our proposed method preserved a substantial amount of visible details with accurate geometry features. The aforementioned quantitative and qualitative results strongly support the effectiveness of our method in achieving high-quality 3D reconstruction scenes at real-time inference speeds. This highlights its potential for future real-time endoscopic applications. We provide more visualizations on StereoMIS in the supplementary.

To further analyze the contributions of our designs, we conducted an ablation study on the EndoNeRF dataset~\cite{wang2022neural} by removing (i) depth regularization, (ii) surface normal constraints, (iii) confidence-guided learning. The experimental results in Table~\ref{tab:ablation} unequivocally demonstrate that the absence of any of the components leads to a substantial degradation in performance. These results highlight the crucial role played by each component in enhancing the quality, accuracy, and overall performance of our method.

\section{Conclusion}
\label{sec:conclusion}

In this paper, we propose Endo-4DGS, a real-time, high-fidelity reconstruction method of deformable tissues. Different from previous works, lightweight MLPs are implemented to capture temporal dynamics with Gaussian deformation fields. We further propose to estimate the depth map by a foundation model Depth-Anything for Gaussian Initialization. The framework is additionally enhanced with confidence-guided strategy, surface normal constraints, and depth regularization to better utilize the depth prior constraint. Extensive experiments demonstrate the superior performance and fast inference speed of our proposed method against other state-of-the-art methods. These results underline the vast potential of Endo-4DGS to improve a variety of surgical applications, allowing for better decision-making and safety during operations.


%
%
%
%

\newpage
\bibliography{reference}{}
\bibliographystyle{splncs04}






\end{document}


\title{Supplementary Materials for ``Endo-4DGS: Endoscopic Monocular Scene Reconstruction with 4D Gaussian Splatting''}
\maketitle
\begin{figure}[h]
    \centering
    \includegraphics[width=0.95\linewidth]{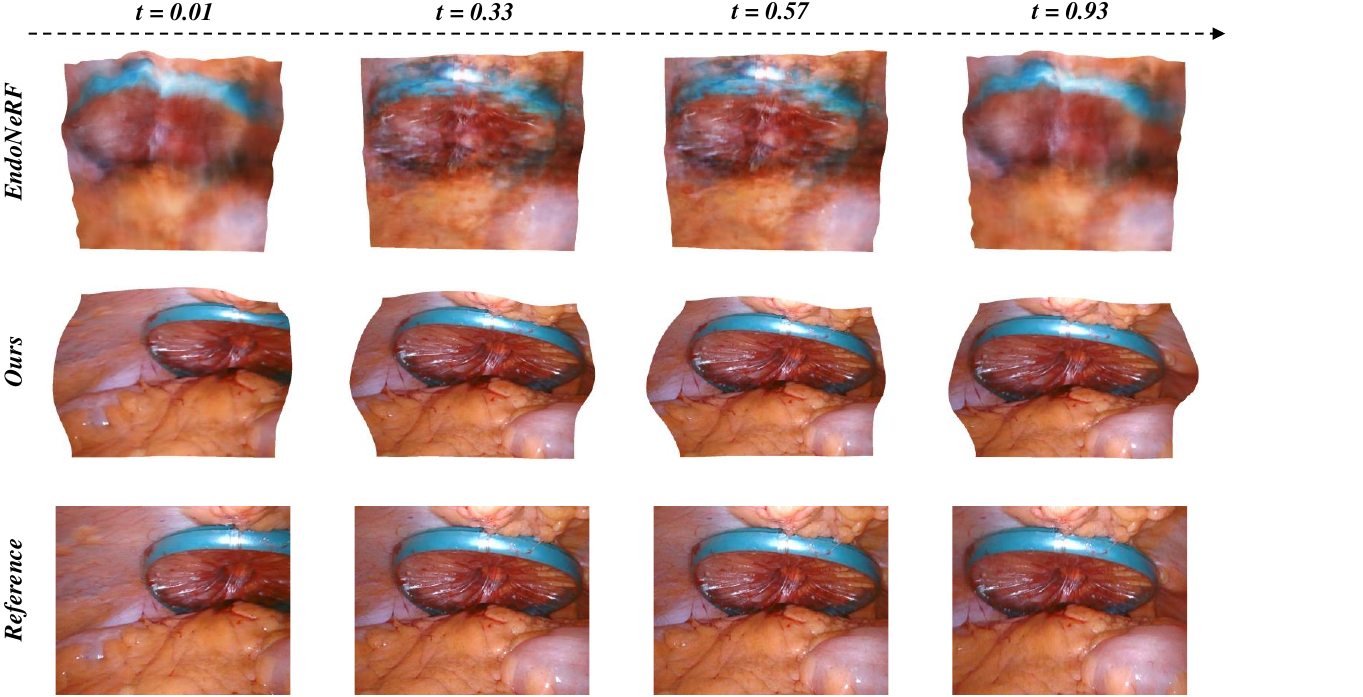}
    \caption{Qualitative comparison on the StereoMIS dataset~\cite{hayoz2023robust} against EndoNeRF~\cite{wang2022neural}.}
    \label{fig:stereomis}
\end{figure}

\begin{table}[h]
\caption{Ablation experiments of the proposed method on EndoNeRF Dataset~\cite{wang2022neural}. We compare the performance by removing the depth initialization.}
\centering
\resizebox{0.9\textwidth}{!}{
\begin{tabular}{c|ccc|ccc}
\noalign{\smallskip}\hline
\multirow{2}{*}{Models}
& \multicolumn{3}{c|}{EndoNeRF-Cutting} & \multicolumn{3}{c}{EndoNeRF-Pulling} \\ \cline{2-7}
& PSNR $\uparrow$ & SSIM $\uparrow$ & LPIPS $\downarrow$ & PSNR $\uparrow$ & SSIM $\uparrow$ & LPIPS $\downarrow$ \\ \hline
Without depth initialization & 5.62 & 0.606 & 0.528 &  7.08 & 0.728 & 0.416 \\
With depth initialization& \textbf{36.56} & \textbf{0.955} & \textbf{0.032} & \textbf{37.85} & \textbf{0.959} & \textbf{0.043}  \\ \hline
\end{tabular}}
\label{tab:ablation_init}
\end{table}

\bibliography{reference}{}
\bibliographystyle{splncs04}